\documentclass[letterpaper, 10 pt, conference]{ieeeconf} 
\IEEEoverridecommandlockouts 
\usepackage{times}  
\usepackage{url}  
\usepackage{xfrac}  
\usepackage{gensymb}
\usepackage{graphicx}  
\usepackage{multirow}
\usepackage{booktabs} 
\usepackage{graphicx}
\usepackage{amssymb}
\usepackage{array}
\usepackage{amsmath}
\usepackage{filecontents,lipsum}
\usepackage{amsmath}
\usepackage{float}
\usepackage{hyperref}
\usepackage{paralist}
\usepackage[linesnumbered,ruled]{algorithm2e}
\title{\LARGE \bf 
Customizing Object Detectors for Indoor Robots
}

\author{Saif Alabachi$^{1}$ \quad Gita Sukthankar$^{2}$ \quad Rahul Sukthankar$^{3}$%
\thanks{$^{1}$Saif Alabachi is with the Department of Computer Engineering, University of Central Florida, Orlando, FL and the University of Technology, Baghdad, Iraq
        {\tt\small s.mohammed@knights.ucf.edu}}%
\thanks{$^{2}$Gita Sukthankar is with the Department of Computer Science, University of Central Florida, Orlando, FL
        {\tt\small gitars@eecs.ucf.edu}}%
\thanks{$^{3}$Rahul Sukthankar is with Google AI
        {\tt\small sukthankar@google.com}}%
}

\begin{document}
\maketitle
\thispagestyle{empty}
\pagestyle{empty}

\begin{abstract}

Object detection models based on convolutional neural networks (CNNs) demonstrate impressive performance when trained on large-scale labeled datasets.  While a generic object detector trained on such a dataset performs adequately in applications where the input data is similar to user photographs, the detector performs poorly on small objects, particularly ones with limited training data or imaged from uncommon viewpoints. Also, a specific room will have many objects that are missed by standard object detectors, frustrating a robot that continually operates in the same indoor environment.

This paper describes a system for rapidly creating customized object detectors.  Data is collected from a quadcopter that is teleoperated with an interactive interface.  Once an object is selected, the quadcopter autonomously photographs the object from multiple viewpoints to 
collect data to train DUNet (Dense Upscaled Network),
our proposed model for learning customized object detectors 
from scratch given limited data.  Our experiments compare the performance of learning models from scratch with DUNet vs.\ fine tuning existing state of the art object detectors, both on our indoor robotics domain and on standard datasets.

\end{abstract}

\section{Introduction}

\begin{figure}
\centering
	\includegraphics[height=10.5cm,width=0.47\textwidth]{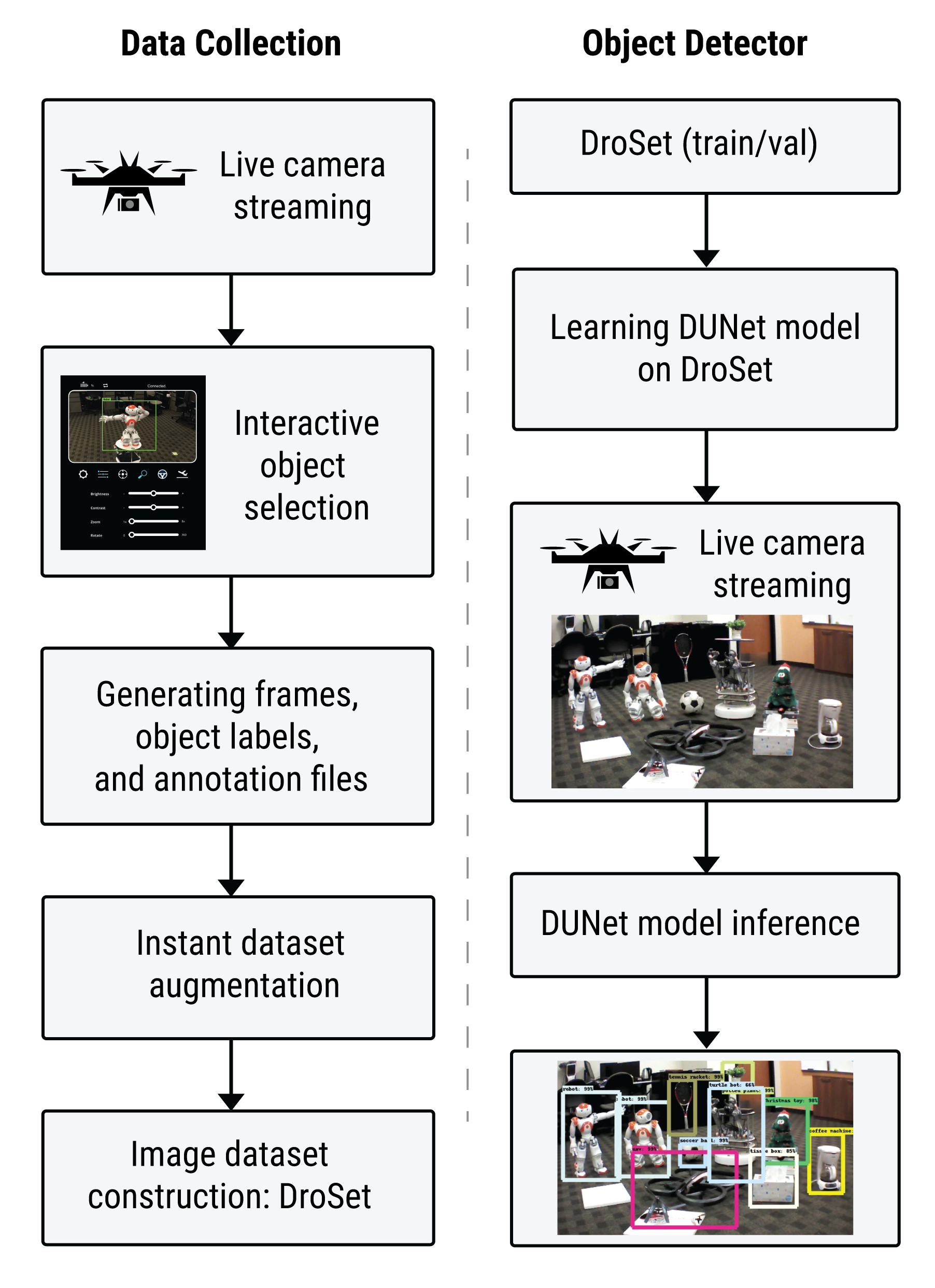}
\caption{Our system consists of two parts: 1) a semi-autonomous data collection system and 2) our neural network architecture for rapidly training custom object detectors.  The user teleoperates the quadcopter toward the object using an interactive user interface.  After the user selects the object, the quadcopter autonomously captures multiple views of the object, which are then augmented with synthetically filtered images.  This dataset (DroSet) was then used to train DUNet (Dense Upscaled Network).  At the conclusion of the procedure, the quadcopter can fly autonomously around the environment rapidly and reliably detecting all objects initially specified by the user.}
\label{fig:platformOverview}
\end{figure}

Indoor mobile robots operate in spaces that are not as rigorously controlled as manufacturing areas, nor as rich with diversity as outdoor scenes.  The open source release of pre-trained object detection models such as the TensorFlow Object Detection API~\cite{huang2017speed} has been a boon to robotics, but in indoor spaces, many objects, particularly small ones, are omitted from the common object datasets.  This is a hindrance for creating indoor robots that can be tasked to find or manipulate objects on tables, walls, and desks.   Our aim is to develop a system that can be used to rapidly create customized detectors for vision-based robots that require real-time object detection.  This is related to the challenge of using deep learning to perform visual SLAM~\cite{wang2017} but with the objective of tasking the robot to use the objects rather than learning landmarks for visual navigation.  This paper addresses the following challenges:
\begin{compactenum}
\item Collecting and annotating images of novel relevant objects with minimal human effort.
\item Developing a CNN architecture that trains with limited data and performs real-time inference on videos.
\end{compactenum}
Figure~\ref{fig:platformOverview} shows our proposed system, which includes an interactive user interface that enables the user to task a quadcopter to autonomously collect images of target objects from multiple viewpoints and label them without additional manual effort.  Quadcopters as imaging platforms have become ubiquitous in recent years in a wide variety of applications including surveillance~\cite{surveillance}, real-estate photography, agricultural and industrial inspection~\cite{sa2014}.  

To address the second challenge, we introduce a new architecture, DUNet (Densely Upscaled Network), that is inspired by the DenseNet~\cite{huang2017densely} image classifier, the feature pyramid network (FPN)~\cite{lin2017feature}, and the meta-architecture of the SSD~\cite{liu2016ssd} object detector.  By combining dense layers and upscaling, DUNet can reliably detect small objects and requires fewer classification layers to achieve the desired speed-quality balance.  

Our data collection platform was used to collect a dataset  (DroSet) of ten real-world objects along with labeling and bounding box annotations. It includes both images captured at different viewpoints and ranges, along with augmented data created by applying filters to create contrast, background, and brightness variations.  A standard approach would be to use DroSet to fine tune an existing network trained on a large dataset such as PascalVOC~\cite{Everingham15}. Fine-tuning can be used to reduce the training time and improve convergence; if the new objects do not share sufficient feature representations with the original dataset then fine tuning performs poorly. In contrast, the dense layers of DUNet promote convergence on small customized datasets. Both our DUNet framework\footnote{Download DUNet from \url{https://github.com/cyberphantom/Customizing-Object-Detectors-for-Indoor-Robots}} and dataset are publicly available.

Our experimental results demonstrate that the DUNet architecture can be trained from scratch on a small dataset, achieves higher accuracy on small-sized objects and 
achieves frame-rate object detection on image streams. DUNet is also practical as a generic object detector, achieving competitive performance on standardized object detection datasets as state of the art models.


\section{Related Work}

Object recognition has a long history in computer vision~\cite{roth2008survey}. The field saw major advances due to the resurgence of neural networks, specifically deep convolutional networks, initially for the task of image labeling~\cite{krizhevsky2012imagenet} and subsequently for detection~\cite{szegedy2014scalable}. The image labeling task (e.g., ImageNet~\cite{deng2009imagenet}) is most relevant to information retrieval and requires assigning to each input (image), a class label corresponding to the dominant semantic category visible in that image. By contrast, object detection or localization, consists of drawing a bounding box around each object (from a set of relevant categories) in the image, along with its semantic label -- and is thus of direct relevance to robotics. We briefly cover both the approaches and relevant datasets below.


Image datasets are typically sourced from the Internet but there is also a growing trend of datasets, particularly for robotics applications, collected directly from the real world. For instance, the KITTI dataset~\cite{Geiger2013IJRR} consists of roadway images taken from a car driving in an urban environment. There has also been significant recent progress in efficiently collecting large quantities of visual data using robots, including smart user interfaces for semi-automated data collection using drones (e.g., \cite{Sukthankar-FLAIRS2018}) and indoor mobile robots (e.g., \cite{loghmani2018}).

Convolutional neural networks (CNNs)~\cite{Lecun1989} were initially applied to handwritten digit recognition but were shown to outperform traditional techniques such as deformable part models~\cite{DPM} on image labeling in AlexNet~\cite{krizhevsky2012imagenet}. Since then, there have been consistent improvements to the state-of-the-art based on extensions to CNN-based architectures, such as VGG~\cite{simonyan2014very}, GoogLeNet~\cite{Szegedy_2015_CVPR} and DenseNet~\cite{huang2017densely}.

CNNs were also instrumental to recent progress on object localization, starting with MultiBox~\cite{szegedy2014scalable}. Inspired by classification models, R-CNN~\cite{girshick2014rich} used cropped boxes from the original image as input to a neural network classifier. Unfortunately, R-CNN was computationally expensive since it repeatedly processed the same pixels whenever they appeared in different overlapping regions. Fast R-CNN~\cite{girshick2015fast} addressed this defect by first pushing the entire image through a feature extractor, thus amortizing the computation across the set of anchor boxes.
This set of ideas has culminated in Faster R-CNN~\cite{faster-r-cnn}, where region proposals are efficiently generated using a fully convolutional network. While Faster R-CNN can process several images per second, it is typically still too slow for most mobile or robotics applications that demand real-time performance on compute-constrained platforms. This has motivated a series of object detection models, such as SSD~\cite{liu2016ssd} and YOLO~\cite{redmon2018yolov3} that aim for high quality detections at near real-time speed.

Our work is informed by the comprehensive experiments on object detection speed/accuracy trade-offs conducted by Huang et al.~\cite{huang2017speed}, where SSD + MobileNet emerges as a very strong baseline for our application. However, we saw opportunities for improving customized object detectors, drawing inspiration from recent work on feature extraction in DenseNet~\cite{huang2017densely}, fully-convolutional approaches to semantic segmentation such as Tiramisu~\cite{tiramisu} and recent multi-scale approaches for object detection, such as FPN~\cite{lin2017feature} and TDM~\cite{tdm}.

The standard approach to customizing an object detector is via domain transfer --- e.g., replacing the final layer in a strong pre-trained model and fine-tuning it on the new data. However, we see significant advantages to training custom object detectors from scratch, such as DSOD~\cite{shen2017dsod}, which demonstrates competitive performance, albeit not in real-time.

Thus, our proposed approach for customizing real-time object detectors, termed Dense Upscaled Network (DUNet), is an architecture inspired by SSD, DSOD, FPN, TDM and is trained from scratch on data collected semi-autonomously by an indoor UAV.

\begin{figure*}[t]
\centering
	\includegraphics[width=0.9\textwidth]{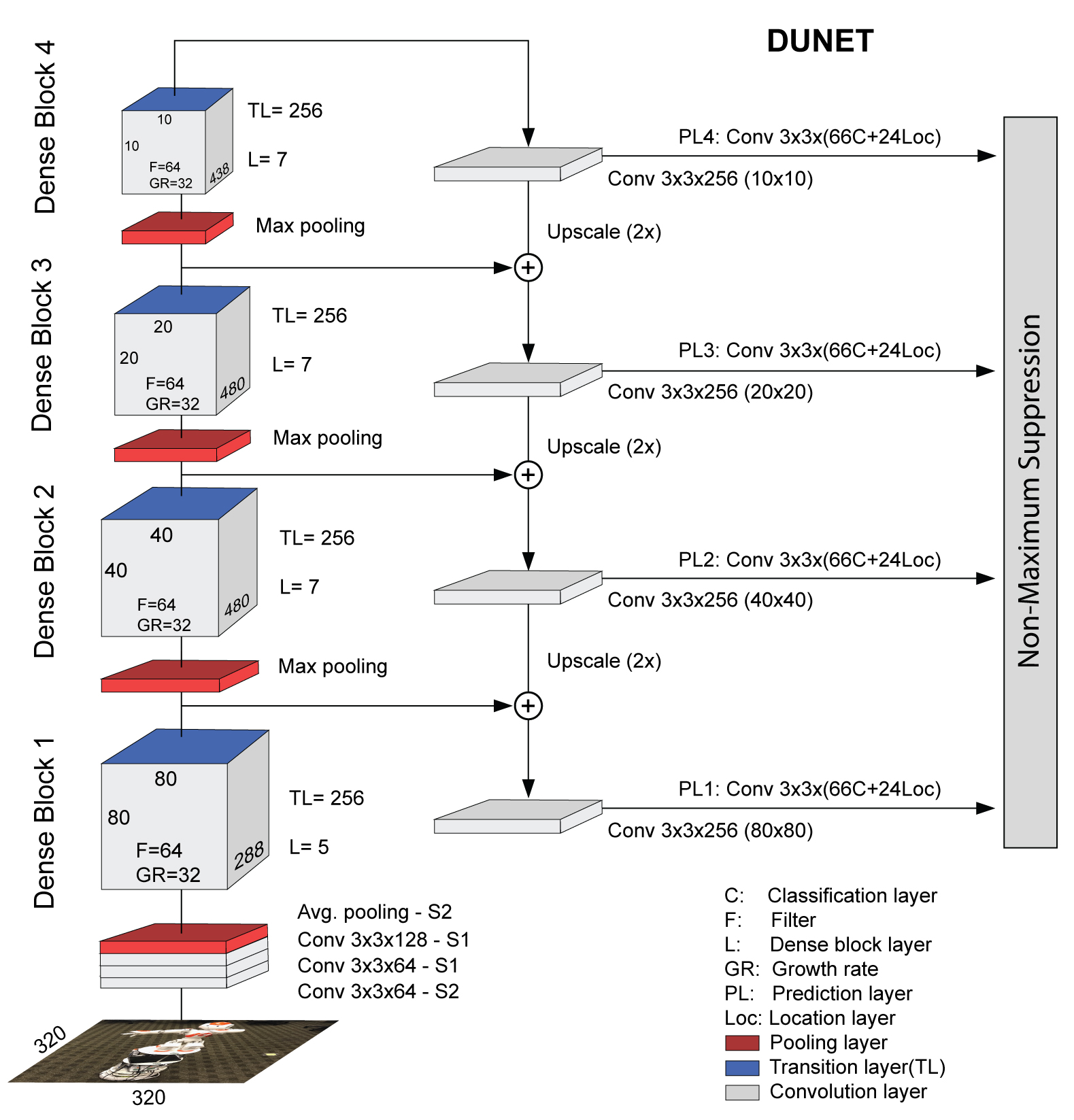}
\caption{DUNet architecture}
\label{fig:DUNet}
\end{figure*}

\section{Dense Upscaled Network (DUNet)}

A typical modern Deep CNN-based object detection system consists of a feature extraction stage combined with an approach for generating bounding box proposals, followed by appropriate classification and regression layers and strategies for non-maxima suppression. For example, an object detector built using the SSD~\cite{liu2016ssd} meta-architecture may employ VGG16~\cite{simonyan2014very} as a feature extraction network, followed by six convolutional classification layers, with four localization points for regressing the ground truth proposals for each class.



The meta-architecture for our model, Dense Upscaled Network (DUNet) is summarized in Fig.~\ref{fig:DUNet}. At a high level, DUNet consists of a sequence of dense blocks that process the input image at different scales, connected to a sequence of prediction layers, each of which independently generate detection results. The two sequences are connected both laterally (at appropriate scales) and vertically (through max pooling and upscaling). We detail each of these aspects below.

\subsection{Feature Extraction}

As discussed above, many object detectors employ a base network for feature extraction; for instance, SSD uses the VGG16 network (pre-trained on ImageNet) for this purpose. In DUNet, we eliminate the use of VGG16 and instead start with a fully convolutional ``initial layer'' sequence followed by average pooling that serves as our feature extractor; this is not pre-trained using ImageNet but is simply randomly initialized and jointly trained from scratch. We employ batch normalization~\cite{batchnorm} before every convolution in DUNet.

DUNet then processes these initial features using a bottom-up pathway of four dense blocks, rather than the ResNet architecture employed by SSD. Like ResNet, dense blocks enable us to avoid the problem of vanishing gradients and we are able to train these for a customized detector from scratch on a relatively small dataset.


The first dense block consists of five layers, while the remaining dense blocks use seven layers each with 64 filters and a growth rate of 32. Each layer of a dense block includes normalization, ReLU and convolution layers, and each layer's input consists of the concatenated outputs of every feature-map from each of the preceding layers.



The top-down pathway (inspired by feature pyramid networks~\cite{lin2017feature} and top-down modulation~\cite{tdm}) consists of prediction layers interspersed with 2$\times$ upscaling operations. The intuition is that this configuration improves detection of small objects based on their context because each of the prediction layers can exploit both high-resolution features and top-down context. Additionally, the lateral connections serve as skip connections that create short pathways from input to output.

To the best of our knowledge, DUNet is the first architecture to exploit both DenseNet-style concatenation (via dense blocks) in the bottom-up pathway and ResNet-style summation (via the upscaling) in the top-down pathway.

\subsection{Meta Architecture Design Choices}

As described above, many aspects of DUNet's design, such as the use of top-down pathway, are motivated by our desire to improve performance on objects that occupy only a small portion of the image. Reliably finding such ``small objects'' is critical for robotics tasks, particularly when navigating a robot towards a semantic landmark that is farther away (e.g., ``go to the fire alarm'').

A straightforward approach towards this goal would have been to add more classification layers to SSD or consider more aspect ratios/scales. However, such an approach would come at significant computational cost. Instead, in DUNet, we are able to \emph{reduce} the number of classification layers from six (SSD) to four, while achieving better performance on small object detection in streaming video.


The trade-off relationship between detection speed and accuracy limits the input size. For instance, in SSD~\cite{liu2016ssd}, the authors demonstrate the difference between two implemented versions of SSD network, SSD300 with 300$\times$300 input size resolution and SSD512 with 512$\times$512. On PascalVOC2007 test, SSD300 has mAP=74.3 and 59 fps, whereas SSD512 has mAP=76.8 and 22 fps, so SSD512 is only 2.5\% better accuracy than SSD300 sacrificing more than 62\% of SSD300 speed. Based on this observation, in DUNet we chose 320$\times$320 as the input size resolution. Our experiments show that this input resolution achieves a good balance between accuracy and speed, given our meta-architecture. As detailed below, DUNet without any pre-training and with random initialization outperforms the most recent state-of-the-art object detection models trained on large datasets like MSCOCO and ImageNet on object detection in streaming video.

\section{Semi-Autonomous Visual Data Collection}

As discussed in the introduction, object detection for indoor robotics imposes different challenges than those encountered in object detection for web imagery, such as the requirements for near real-time processing of input video streams, the importance of reliably detecting small-sized objects and the ability to customize the detector for new object classes from limited labeled data. Fortunately, we can also benefit from several features of indoor environments, such as limited variability in terms of lighting, viewpoint, range and background conditions. Here, we present an approach for acquiring training data with minimal human labeling as well as a public dataset (DroSet) for evaluating object detectors on streaming video in such environments.

\subsection{Background}

Collecting labeled datasets for object detection (e.g., PASCAL VOC~\cite{Everingham15}, COCO~\cite{lin2014microsoft}) is significantly more onerous than labeling datasets for whole-image classification (e.g., ImageNet~\cite{deng2009imagenet}). This is because each instance of a relevant object in the image must be localized using a bounding box, which can take several seconds per instance even for an expert annotator.

When the input images consist of a video stream, manually labeling each frame becomes impractical and it is important to consider semi-automated schemes for labeling that exploit temporal consistency.


\subsection{Interactive Data Collection}

We collect training and evaluation data using an indoor drone and a semi-autonomous user interface (SUI)~\cite{Sukthankar-FLAIRS2018}. The user interactively selects objects of interest and the tracking agent controls the drone to collect a stream of images capturing the object from multiple viewpoints, by tracking the object while flying in a variety of patterns. The system minimizes annotator effort by exploiting temporal consistency since the tracker automatically propagates the bounding box around the object from frame to frame. This data is then used to train DUNet and enables repeatable object detection experiments.


\subsection{Live Data Augmentation}
Synthetic image augmentation~\cite{pomerleau1992} is performed on a captured image using a series of 2D geometric transforms (e.g., rotation, translations) and induced photometric variations (e.g., brightness, contrast and color shifts). The resulting set of images for each object instance are much richer than those that would be typically obtained from the Internet since they include variations in appearance induced by viewpoint changes as well as specular reflections from changing lighting (relative to camera).

Given that there is significant redundancy across consecutive images in the image stream, we choose a slightly different data augmentation strategy than is commonly employed on standard image datasets.
Rather than applying all of the augmentation filters on each image, our system captures a fresh frame before applying each filter (to further introduce slight variations). Thus, successively captured frames are processed by each of the transformations. 

We include all of the common geometric and photometric transformations, such as brightness, contrast, rotation, flipping, shadow, background, and color shift. The user can interactively add or remove filters as desired during the capture process, as well as selecting the rate at which each filter is applied (e.g., if more rotations vs.\ contrast changes are desired). Since each filter is applied to a freshly captured frame, the data generator generates fewer ``near-duplicate'' instances in its dataset than traditional data augmentation schemes that are forced to apply all filters on each original image. We specify 500~ms as a minimum threshold between consecutively captured images in order to allow sufficient time for the drone to change its position (gaining more variations in depth and angle view point). 
%
Our data collection system records input at 640$\times$360 resolution at a rate of $\sim$30.6 frames/second.

\subsection{DroSet: A Dataset of Indoor Objects}

Following the procedure described above, we captured footage of 10 object categories in indoor environments and organized it into training (75\%), validation (15\%) and test (10\%)  sets. This dataset, termed DroSet, has been released publicly\footnote{The DroSet dataset is available at \url{https://goo.gl/xE6Jkr}} to enable other researchers to evaluate their object detection algorithms under our conditions.

DroSet consists of image streams for the following ten  categories of indoor objects: christmas toy, coffee machine, potted plant, tissue box, robot, soccer ball, turtle bot, UAV, fire alarm, and tennis racket. By design, three of these categories (e.g., potted plant) overlap with categories in COCO, while the others are new. Some of the object categories exhibit little visual variation (e.g., fire alarm), while others (e.g., UAV) contain objects with very different appearances.
Our choice of categories should enable researchers to better evaluate the extent to which transfer learning generalizes from standard datasets to our dataset for both the overlapping and new categories.




\section{Experimental Results}

For indoor robotics applications, it is important that proposed methods find a good balance between processing speed and detection quality. Thus, our primary experimental scenario (Sec.~\ref{sec:scenario1}) evaluates methods on a 30fps input stream of frames. However, it is valuable to confirm that our proposed model is competitive on traditional object detection metrics, so we also include a direct comparison of DUNet against SSD on a standard dataset (Sec.~\ref{sec:scenario2}).

DUNet is implemented using Keras with the TensorFlow~\cite{abadi2016tensorflow} back-end. We use the TensorFlow Object Detection API~\cite{huang2017speed} implementations for all of the baseline models, such as SSD-300. All of the experiments were conducted on a machine with an NVIDIA GeForce GTX Titan X graphics card.

\subsection{Scenario I: Evaluation on Real-Time Robot Input Stream}
\label{sec:scenario1}

We use the Robotics Operating System (ROS)~\cite{quigley2009ros} to record frame streams captured by the quadcopter camera. This enables us to create repeatable playback environments for testing the different models under realistic robot conditions. For this scenario, we create bag test files for each of the ten DroSet categories (where exactly one instance of the given object appears in each frame) to enable computation of per-class results. These are available in the public DroSet release.

\begin{figure}
\centering
	\includegraphics[width=1.0\columnwidth]{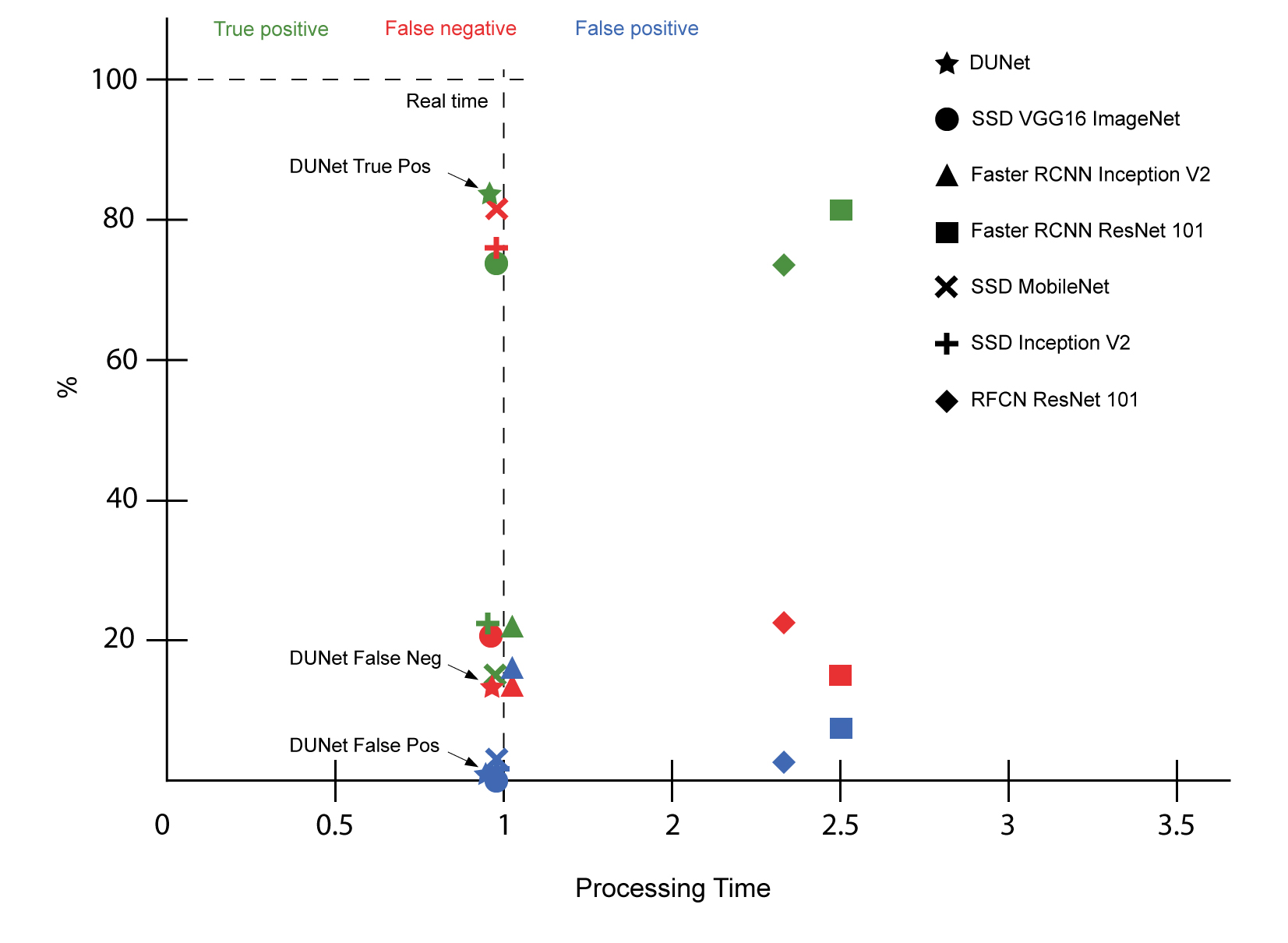}
\caption{
Scatterplot of detector quality (TP, FN, FP) on DroSet vs.\ processing time (normalized to real-time) for each model. DUNet clearly outperforms other models while processing input stream in real time.
}
\label{fig:time}
\end{figure}

Fig.~\ref{fig:time} compares DUNet against a comprehensive array of state-of-the-art models, both in terms of detection quality (true positive, false negative and false positive rates) and processing time. We observe that DUNet clearly outperforms real-time baselines like SSD VGG16 ImageNet in terms of detection quality, and is 2.5$\times$ faster than state-of-the-art models like Faster R-CNN + ResNet 101, which are unable to keep up with the input stream. These results on DroSet are consistent with the speed/accuracy experiments reported on standard datasets~\cite{huang2017speed}.

\begin{figure}
\centering
\includegraphics[width=1.0\columnwidth]{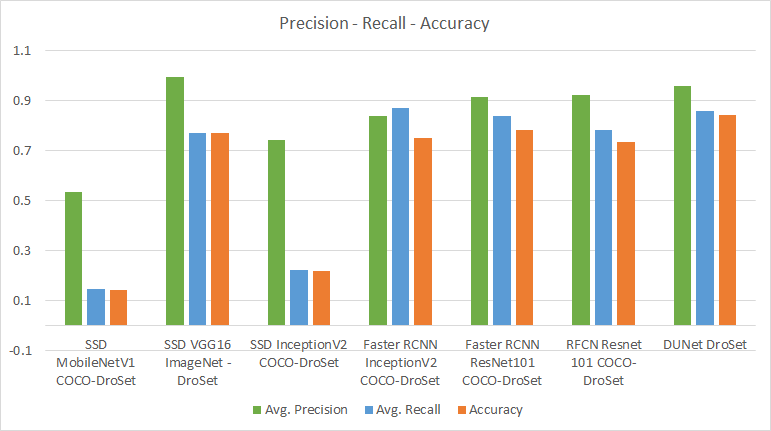}
\caption{Average precision, recall, and accuracy on DroSet for the fine-tuned, pre-trained baseline models vs.\ DuNet (trained from scratch).}
\label{fig:pr_rc_ac}
\end{figure}

Fig.~\ref{fig:pr_rc_ac} summarizes overall average precision, recall and accuracy on DroSet for all of the models and DUNet is the clear winner.

\begin{figure*}
\centering
	\includegraphics[height=7.0cm,width=1.0\textwidth]{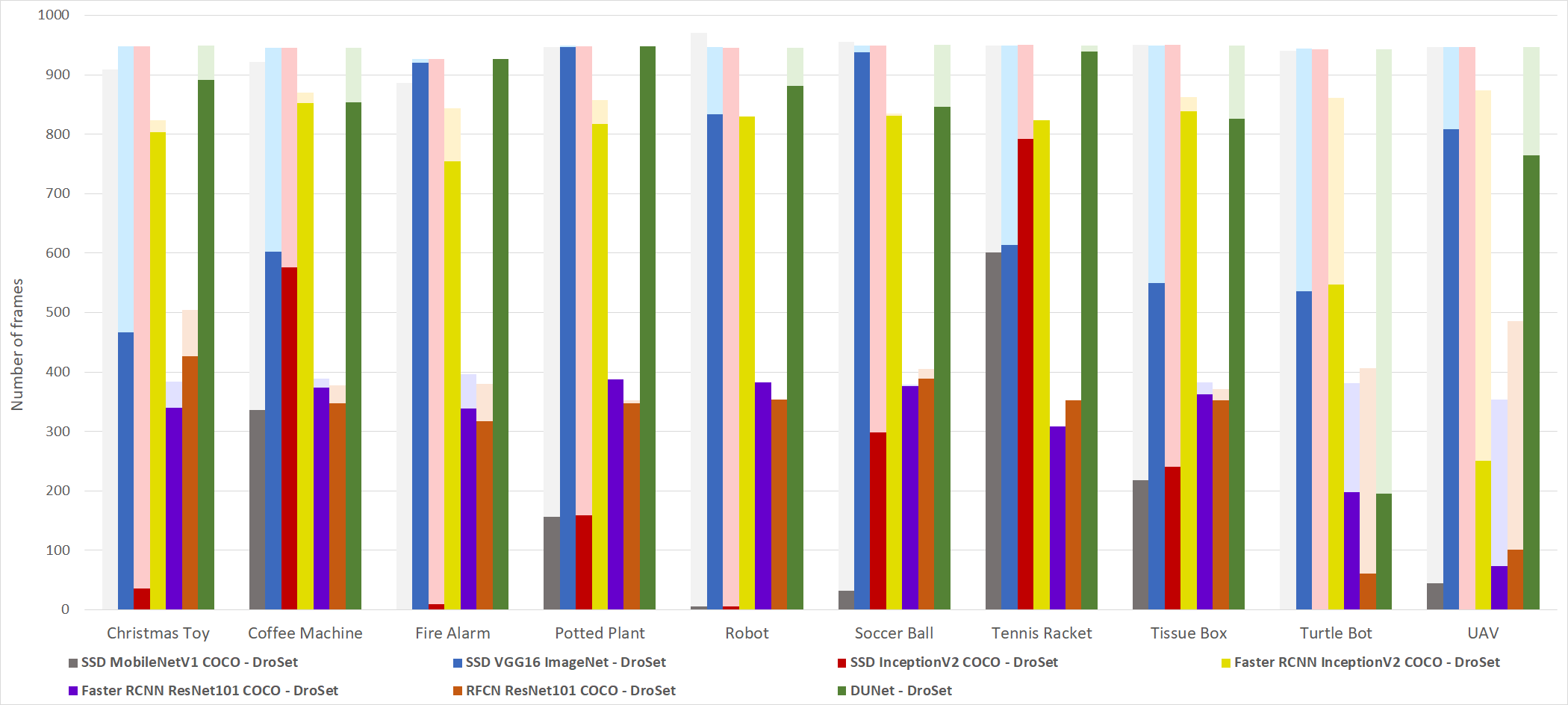}
\caption{
Number of processed frames and recall for each model, per DroSet category. Each column corresponds to a model, with translucent bars showing the number of processed frames and dark bars denoting the correct detections. DUNet and SSD-based models process more frames than others, and DUNet (far right column in each set) has the highest recall in almost every category. Note that many state-of-the-art models perform poorly on several categories.
}
\label{fig:allobjects}
\end{figure*}

Fig.~\ref{fig:allobjects} presents a more detailed breakdown of each model (shown as a column) on the subset of sequences featuring a given DroSet category. The translucent bars correspond to the number of frames processed by each model in steady state. Most models either fail to process sufficient frames or exhibit low detection rate. We also see that fine-tuning standard pre-trained detectors on DroNet can vary widely: e.g., SSD MobileNet V1 pre-trained on COCO and fine-tuned on DroSet does well on tennis racket but terribly on christmas toy. Interestingly, there is not a clear correlation between the domain transfer performance for such baseline models and categories that overlap with COCO vs.\ new categories. DUNet (despite being trained from scratch) wins on both metrics on almost all of the classes.

\begin{table*}
\caption{
Direct comparison to SSD on PASCAL VOC (following protocol for Table~1 in \cite{liu2016ssd} with union of PASCAL VOC 2007+2012 train/val) confirming that the proposed DUNet model is competitive on standard object detection benchmarks.
}
\label{table:voc}
\renewcommand*{\arraystretch}{1.4}
\centering
\begin{tabular}{ p{1.0cm}|p{0.5cm}|p{0.35cm}p{0.35cm}p{0.35cm}p{0.35cm}p{0.35cm}p{0.35cm}p{0.35cm}p{0.35cm}p{0.35cm}p{0.35cm}p{0.35cm}p{0.35cm}p{0.35cm}p{0.35cm}p{0.35cm}p{0.35cm}p{0.36cm}p{0.35cm}p{0.35cm}p{0.35cm}}
Model & mAP & aero & bike & bird & boat & botle & bus & car & cat & chair & cow & table & dog & horse & mbik & persn & plant & sheep & sofa & train & tv \\
\hline
SSD300       & 74.3	& 75.5 	& 80.2 	& 72.3 	& 66.3 	& 47.6 	& 83.0 	& 84.2 	& 86.1 	& 54.7 	& 78.3 	& 73.9 	& 84.5 	& 85.3 	& 82.6 	& 76.2 	& 48.6 	& 73.9 	& 76.0 	& 83.4 	& 74.0\\ 
DUNet   	& 74.3	& \textbf{83.0}	& \textbf{82.3}	& 69.5	& 62.5	& 37.7	& \textbf{85.0}	& \textbf{88.0}	& 84.2	& \textbf{56.2}	& 76.1	& 73.6	& 80.4	& \textbf{87.8}	& 82.5	& \textbf{79.8}	& 46.1	& \textbf{76.3}	& 75.0	& \textbf{84.9}	& \textbf{75.4} \\ 
\noalign{\vskip 2mm} 
\end{tabular}
\end{table*}

\subsection{Scenario~II: Evaluating DUNet on Standard Benchmark}
\label{sec:scenario2}

The second scenario evaluates DUNet under standard object detection conditions on traditional object detection benchmarks, against state-of-the-art models. This is primarily to confirm that our proposed meta-architecture is indeed competitive under such conditions and not overly specialized for our use case.

For this evaluation, we trained a DUNet model from scratch on PASCAL VOC~\cite{Everingham15}, with the final layer replaced with PASCAL VOC object categories. We chose SSD300 as a strong baseline based on speed/accuracy results reported in Huang et al.~\cite{huang2017speed} and replicated the experimental methodology described in the SSD paper~\cite{liu2016ssd}.

Table~\ref{table:voc} presents a direct comparison of SSD vs.\ DUNet. The dataset was the union of PASCAL VOC07 and VOC12, with results for SSD300 (first row) copied directly from the SSD paper~\cite{liu2016ssd}. We see that DUNet trained from scratch performs as well as SSD300, which includes a VGG16 feature extractor trained on ImageNet. Note that we did not optimize the DUNet performance on VOC for this scenario (e.g., through hyperparameter tuning).

In summary, our experiments show the effectiveness of DUNet, both in its primary role as a strong meta-architecture for training customized real-time object detectors for indoor robots, as well as its competitiveness in standard conditions.

\section{Conclusion}

The paper introduces DUNet, a novel meta-architecture for real-time object detection. Our design choices focus on reliable detection of small-sized objects through the use of dense blocks and top-down context, as well as customization of detectors for new object classes via training from scratch on limited datasets. We have made the data used for our evaluation publicly available----DroSet, a dataset of indoor objects, collected semi-autonomously using a drone. This dataset consists of frame streams that can be played back in a repeatable manner so as to evaluate object detectors in robotics applications. 

Our experiments confirm that DUNet outperforms current state-of-the-art models on real-time object detection for indoor robotics. Additionally, even when trained from scratch, DUNet is competitive on standard object detection benchmarks.

Our DUNet implementation and the DroSet dataset have been made publicly available to encourage further research in this area.

\section*{Acknowledgments}

The authors would like to thank Yasmeen Alhamdan for help in generating the DUNet network architecture diagrams and figures for experimental results.

\bibliography{bib}

\end{document}